\documentclass[conference]{IEEEtran}
\IEEEoverridecommandlockouts
\usepackage{cite}
\usepackage{amsmath,amssymb,amsfonts}
\usepackage{algorithmic}
\usepackage{graphicx}
\usepackage{textcomp}
\usepackage{xcolor}
\usepackage{array}
\usepackage{booktabs}
\usepackage{multirow}
\usepackage{ textcomp }
\usepackage{placeins}
\begin{document}

\title{SCINTILLA-SNN: A Spiking Multi-Scale Selective Aggregation Network for Perineural Invasion Prediction}


\author{
\IEEEauthorblockN{
Youngung Han$^{1,2,*}$, Yului Jeong$^{1,*}$, Kyeonghun Kim$^{2}$, Dohyun Kweon$^{2,3}$, Suah Park$^{1}$, \\
Hyunsu Go$^{1}$, Sungha Park$^{1,4}$, Anna Jung$^{1}$, Jinyong Jun$^{1}$, Yunho Choe$^{1}$, Yunjin Seo$^{1}$,\\ Ken Ying-Kai Liao$^{5}$,
Hyuk-Jae Lee$^{1}$, Nam-Joon Kim$^{1,\dagger}$
}

\IEEEauthorblockA{\footnotesize
$^{1}$Seoul National University, Seoul, Republic of Korea \quad
$^{2}$OUTTA, Seoul, Republic of Korea
}

\IEEEauthorblockA{\footnotesize
$^{3}$Kyung Hee University, Seoul, Republic of Korea \quad
$^{4}$Seoul National University School of Medicine, Seoul, Republic of Korea
}

\IEEEauthorblockA{\footnotesize
$^{5}$NVIDIA AI Technology Center, Taipei, Taiwan
}

\IEEEauthorblockA{\footnotesize
$^{*}$These authors contributed equally to this work. \\
$^{\dagger}$Corresponding author: \texttt{knj01@snu.ac.kr}
}
}

\maketitle

\begin{abstract}
Preoperative prediction of perineural invasion (PNI) in cholangiocarcinoma (CCA) is clinically valuable but remains challenging because PNI-related cues on magnetic resonance imaging (MRI) are subtle, sparse, and spatially localized around the tumor boundary. Standard 3D CNN and transformer architectures process volumetric data in a dense or spatially uniform manner, which can dilute subtle PNI-related evidence while requiring a large number of multiply-accumulate operations over 3D feature grids. 
To address these limitations, we propose SCINTILLA-SNN, a 3D spiking network composed of a four-stage hierarchical backbone and a Multi-Scale Spike Aggregation (MSSA) module for PNI prediction. The backbone extracts hierarchical volumetric representations through spiking convolutional stages and local spike window modulation stages. Given the resulting stage-wise representations, MSSA maps each spatial token to a learnable content value and modulates it with a spike-dynamics gate derived from firing rate and timestep-wise membrane-potential variability. The resulting score, referred to as the diagnostic token score, is used to selectively aggregate sparse PNI-related evidence.
Experiments on a 10-year retrospective cohort of 182 CCA patients show that SCINTILLA-SNN achieves an AUROC of 0.748 under 5-fold cross-validation, while reducing the estimated inference energy by 23.18$\times$ compared with dense MAC-only computation of the same network.
\end{abstract}

\begin{figure*}[!t]
\centering
\includegraphics[width=0.95\textwidth]{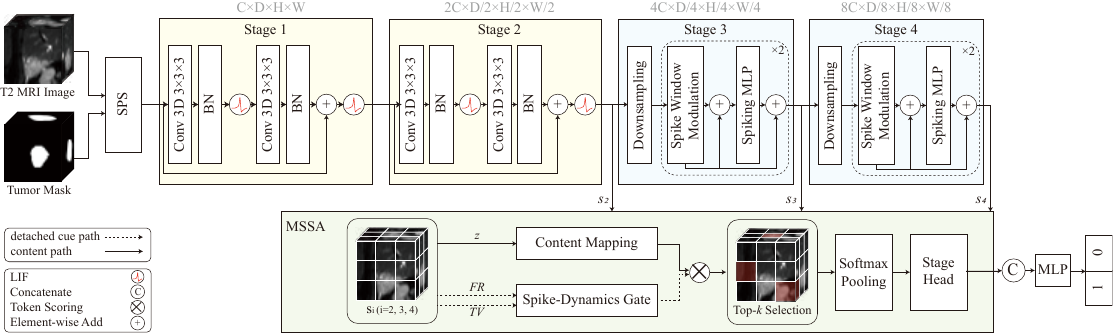}
\caption{Overall architecture of SCINTILLA-SNN. A two-channel T2W volume
is processed over $T=4$ timesteps by an SPS stem and a four-stage
hierarchical spiking backbone, where Stages~1--2 use residual spiking
convolution blocks and Stages~3--4 use spike window modulation blocks.
MSSA aggregates the multi-scale outputs $s_2$, $s_3$, and $s_4$ using
content mapping, spike-dynamics cues, and top-$k$ selection for PNI
classification.}
\label{fig:arch}
\end{figure*}

\section{Introduction}

Perineural invasion (PNI) is an important pathological indicator in cholangiocarcinoma (CCA)~\cite{liebig2009perineural}. Since PNI status is usually confirmed only after surgical resection, reliable preoperative prediction from MRI is clinically meaningful~\cite{conti2026perineural,li2020perineural,wei2022prognostic}. However, MRI-based PNI prediction remains challenging because PNI-related findings are often subtle, sparse, and localized near the tumor periphery. Image-based PNI prediction has been investigated using radiomics, clinical variables, and deep-learning models. Radiomic approaches rely on predefined handcrafted features and typically require explicit feature selection to address the high dimensionality of the extracted feature space~\cite{lambin2017radiomics, van2017computational}. More recently, 3D CNN and transformer-based models learn volumetric representations directly from MRI~\cite{han2026losa, han2026mma, han2026neonet, um2026spikeds, han2026adaptive}.

However, many deep models process volumetric feature grids in a dense or spatially uniform manner~\cite{dosovitskiy2020image, rao2021dynamicvit}, which may weaken subtle and localized PNI-related signals. Moreover, dense processing of 3D MRI requires a large number of multiply-accumulate (MAC) operations over volumetric feature grids~\cite{liu2021swin, dosovitskiy2020image}, increasing inference energy consumption and making such models less practical for resource-constrained clinical settings.

Spiking neural networks (SNNs) offer a promising direction for
energy-efficient inference by replacing multiply-accumulate (MAC)
operations with accumulate (AC) operations in the sparse spike
domain~\cite{roy2019towards, yin2021accurate}. Recent spiking transformers extend SNNs with spike-based attention mechanisms to improve representational capacity, but mainly address efficient feature computation at the architecture and attention level~\cite{zhou2022spikformer, yao2023spike, wang2023masked, zhou2024qkformer, zhou2023spikingformer}. In MRI-based PNI prediction, however, the challenge is not only efficient computation
but also preserving subtle and spatially localized diagnostic signals
within redundant 3D volumes. Spatially uniform aggregation may dilute these sparse signals, motivating a 3D spiking framework that combines sparse spike-based computation with selective, diagnosis-oriented aggregation of PNI-relevant volumetric patterns~\cite{kim20263d, han2025foscu, kim2026maesil}.

To this end, we propose SCINTILLA-SNN, a hierarchical spiking network for MRI-based PNI prediction. The model combines a four-stage spiking backbone with a Multi-Scale Spike Aggregation (MSSA) module, which modulates learnable content values with spike-dynamics cues to selectively aggregate diagnostic evidence across multiple feature scales. 

The main contributions of this work are as follows. First, we propose MSSA, a multi-scale spike aggregation module that selectively aggregates informative tokens across multiple backbone stages, replacing spatially uniform pooling of volumetric spike features. Second, we introduce a spike-dynamics gate that combines firing-rate activity and timestep-wise membrane-potential variability with learnable content values, enabling token selection based on both feature
content and spike-dynamics cues. Third, we validate SCINTILLA-SNN on a retrospective cohort of 182 cholangiocarcinoma patients, where it achieves the highest AUROC among the evaluated baselines. Ablation studies further support the contribution of spike-dynamics gating and selective token aggregation, while operation-level energy analysis provides an additional view of its efficiency.
\section{Method}
\label{sec:method}

\subsection{Overview}

SCINTILLA-SNN is a 3D spiking neural network (SNN) that maps a two-channel input volume $x \in \mathbb{R}^{B \times 2 \times D \times H \times W}$ to binary PNI logits. The two channels are the normalized T2-weighted (T2W) intensity image and the tumor mask. As shown in Fig.~\ref{fig:arch}, the input volume is processed through (i) a 3D spiking patch stem (SPS), (ii) a four-stage hierarchical spiking backbone, (iii) a multi-scale spike aggregation (MSSA) module operating on the outputs of Stages~2--4, denoted as $s_2$, $s_3$, and $s_4$, and (iv) an MLP classifier.

\subsection{LIF Spiking Dynamics}

Under direct encoding, the input volume $x$ is fed to the network for
$T$ timesteps. We adopt a standard LIF neuron with hard
reset~\cite{Wu_2018, neftci2019surrogate}:
\begin{equation}
\tilde{u}_t=\beta u_{t-1}+I_t,\quad
s_t=\mathrm{H}(\tilde{u}_t-v_\mathrm{th}),\quad
u_t=\tilde{u}_t(1-s_t)
\label{eq:lif}
\end{equation}
where $\beta$ is the membrane decay factor, $v_\mathrm{th}$ is the firing threshold, and $\mathrm{H}(\cdot)$ is the Heaviside step function~\cite{neftci2019surrogate}. Here $\tilde{u}_t$ denotes the pre-reset
membrane potential, which is used in MSSA to compute membrane-potential
variability.

\subsection{3D Hierarchical Spiking Backbone}

The backbone uses a base channel width of $C{=}32$ with channel multipliers $\{1,2,4,8\}$ across the four stages. The SPS projects the 2-channel input to $C$ channels using two Conv3D--BN--LIF blocks with stride 1. Stage~1 uses a stride-1 spiking residual convolution block: Conv3D--BN--LIF
is followed by Conv3D--BN, shortcut addition, and a final LIF activation. Stage~2 follows the same residual formulation, but uses stride~2 in the
first convolution and a projection shortcut for downsampling, producing
the output $s_2$. Stages~3 and 4 each begin with a single stride-2 spiking downsampling
operation, followed by two spike window blocks. Each block contains a
window-modulation residual sub-block and a spiking-MLP residual
sub-block, summarized as BN--spike window modulation--Add--LIF and
BN--spiking MLP--Add--LIF, respectively. These stages produce $s_3$
with $4C$ channels and $s_4$ with $8C$ channels. The multi-scale outputs $s_2$, $s_3$, and $s_4$ are passed to MSSA.

\subsection{Spike Window Modulation}

Stages~3 and 4 operate on local 3D windows. Within each window,
convolutional projections followed by LIF neurons generate spike-form
$Q$, $K$, and $V$. We then apply a lightweight element-wise modulation:
\begin{equation}
  Y = \eta\bigl((Q \odot K) \odot V\bigr)
  \label{eq:spike_window_modulation}
\end{equation}
where $\odot$ denotes element-wise multiplication,
$\eta=1/\sqrt{d_h}$ is a fixed scaling factor,
and $d_h$ denotes the per-head channel dimension. This operation provides local spike-form feature modulation within each 3D window without dense softmax
attention. The modulated features are projected back to the stage
feature space and combined with the block input through residual
addition followed by LIF activation. The subsequent spiking MLP is implemented with point-wise Conv3D--BN--LIF--Dropout--Conv3D--BN layers, followed by residual addition and a final LIF activation.

\subsection{Multi-Scale Spike Aggregation}
\label{sec:mssa}

MSSA is the main aggregation module of SCINTILLA-SNN. Given the stage
outputs $s_2$, $s_3$, and $s_4$, MSSA assigns a content value
to each spatial token, which is then modulated into a diagnostic token score
used to aggregate only the most informative tokens from each stage. For
each backbone stage $i \in \{2,3,4\}$, MSSA maintains one content path
and two detached spike-dynamics cue paths. The content path
$\bar{f}_i \in \mathbb{R}^{B \times C_i \times D_i \times H_i \times W_i}$
is obtained by averaging the per-timestep stage outputs over $t$ and is
spatially flattened into token features
$z_i \in \mathbb{R}^{B \times N_i \times C_i}$, with $N_i{=}D_iH_iW_i$.
The detached cue paths consist of the binary spike-output stack $S_i$
and the pre-reset membrane-potential stack $M_i$, both in
$\mathbb{R}^{T \times B \times C_i \times D_i \times H_i \times W_i}$,
flattened along the same dimensions to yield per-token firing
rate (FR) and timestep-wise membrane-potential variability (TV).

\textbf{Content mapping.}
A two-layer MLP with LayerNorm and GELU maps each token to a learnable
scalar content value:
\begin{equation}
  \phi_i\bigl(z_i(n)\bigr) \in \mathbb{R}
  \label{eq:phi}
\end{equation}

\textbf{Spike-dynamics cues.}
MSSA computes two scalar spike-dynamics cues for each token: FR from the
spike-output stack and TV from the pre-reset membrane-potential stack,
\begin{align}
  \mathrm{FR}_i(n) &= \mathrm{mean}_t\,\mathrm{mean}_c\,
                      S_{i,t,c,n} ,\label{eq:fr}\\
  \mathrm{TV}_i(n) &= \mathrm{mean}_c\,\mathrm{std}_t\,
                      M_{i,t,c,n} \label{eq:tv}
\end{align}
TV is computed from the pre-reset membrane-potential stack rather than
from binary spikes to capture real-valued membrane fluctuations that are
complementary to FR. Before gating, the content value, FR, and TV are z-score
normalized across spatial tokens within each stage to reduce scale
mismatch:
\begin{equation}
  \widehat{x}_i(n)
  = \frac{x_i(n) - \mu_n(x_i)}{\sigma_n(x_i) + \epsilon},
  \quad
  x_i \in \{\phi_i, \mathrm{FR}_i, \mathrm{TV}_i\}
  \label{eq:zscore}
\end{equation}

\textbf{Diagnostic token score.}
MSSA uses a spike-dynamics gate to modulate the normalized content value
into a diagnostic token score:
\begin{equation}
  g_i(n) = \sigma\!\left(
    \mathbf{w}_i^\top
    \bigl[
      \lambda_\mathrm{FR}\,\widehat{\mathrm{FR}}_i(n),\;
      \lambda_\mathrm{TV}\,\widehat{\mathrm{TV}}_i(n)
    \bigr]
    + b_i
  \right)
  \label{eq:gate}
\end{equation}
\begin{equation}
  \mathrm{score}_i(n)
  = \mathrm{softplus}\!\left(\widehat{\phi}_i(n)\right) \cdot g_i(n)
  \label{eq:score}
\end{equation}
where $\sigma(\cdot)$ is the sigmoid function, and $\mathbf{w}_i$ and 
$b_i$ are stage-specific learnable gate parameters shared across
tokens within stage~$i$. The scalars $\lambda_\mathrm{FR}$ and $\lambda_\mathrm{TV}$ control the
contribution of the FR and TV cues.

\textbf{Top-$k$ pooling.}
For each stage, we retain the $k_i$ tokens with the largest diagnostic
scores and aggregate them using a score-weighted softmax:
\begin{equation}
p_i =
\sum_{n \in \mathcal{T}_i^{(k)}}
\mathrm{softmax}_n\!\big(\mathrm{score}_i(n)\big)\, z_i(n)
\label{eq:pool}
\end{equation}
The pooled vector is projected by a per-stage head $h_i(\cdot)$,
implemented as LayerNorm--Linear--GELU--Dropout, and the
resulting multi-scale evidence vectors are concatenated and
normalized:
\begin{equation}
e_i = h_i(p_i) \in \mathbb{R}^{B \times d_e},
\quad
e = \mathrm{LN}([e_2; e_3; e_4])
\label{eq:evidence}
\end{equation}
where $e \in \mathbb{R}^{B \times 3d_e}$.

\subsection{Objective Function}

The classifier is an MLP head that produces two-class logits. The training objective combines a class-weighted focal loss~\cite{lin2017focal} with label smoothing and two lightweight regularizers,
\begin{equation}
\mathcal{L} = \mathcal{L}_\mathrm{focal}
+ \lambda_\mathrm{fire}\,\mathcal{R}_\mathrm{fire}
+ \lambda_\mathrm{lasso}\,\mathcal{R}_{w}
\label{eq:loss}
\end{equation}
where $\mathcal{R}_\mathrm{fire}$ is an $L_1$ penalty on the mean firing activity computed from the stage spike-output stacks $S_2$, $S_3$, and $S_4$, encouraging sparse firing, and $\mathcal{R}_w$ is a standard  $L_1$ weight penalty.

\section{Experiments}
\label{sec:exp}

\subsection{Dataset and Implementation Details}

\textbf{Dataset.}
We used T2-weighted MRI from a private single-center retrospective cohort
collected over more than a decade, consisting of 182 patients with
histologically confirmed cholangiocarcinoma. Of these patients, 112 were PNI-negative and 70 were PNI-positive, corresponding to a positive-class proportion of 38.5\%. Tumor and liver segmentation masks were manually delineated by radiology experts using 3D Slicer~\cite{fedorov2012slicer}.
Each volume was resampled to a voxel spacing of
$1.5625{\times}1.5625{\times}4.0$~mm and cropped into a tumor-centered
region of $96{\times}96{\times}48$ voxels~\cite{liu2024noninvasive}. A two-channel input was constructed by channel-wise stacking the normalized intensity image and the tumor mask; the intensity channel was processed with percentile clipping followed by z-score normalization, while the tumor mask was kept as a binary float channel. We performed patient-level 5-fold cross-validation and selected checkpoints based on validation AUROC. All baselines were trained and evaluated using the same two-channel input protocol.

\textbf{Implementation.}
Models were trained for 100 epochs per fold with batch size 2 on a single
NVIDIA A100 GPU using AdamW~\cite{loshchilov2019decoupled} with a learning rate of $10^{-4}$, weight decay
of $10^{-4}$, cosine annealing, gradient clipping at 1.0, and mixed-precision
training. Data augmentation was limited to axial flips, applied identically to
the image and mask channels, and intensity jitter applied only to the image
channel.

SCINTILLA-SNN used base channel width of $C{=}32$, $T{=}4$ LIF timesteps, dropout
0.3, two spike window blocks in each of Stages~3 and 4, four Q/K/V
heads, an MLP ratio of 2.0, and local window sizes of
$3{\times}4{\times}4$ for Stage~3 and $3{\times}3{\times}3$ for Stage~4.
The MSSA evidence dimension was set to $d_e{=}128$, with top-$k$ values of
512, 256, and 128 for $s_2$ ($N_2{=}55{,}296$; $\approx 0.93\%$), $s_3$
($N_3{=}6{,}912$; $\approx 3.70\%$), and $s_4$ ($N_4{=}864$;
$\approx 14.81\%$), respectively. Because the content value, FR, and TV are z-score normalized across
spatial tokens within each stage before gating, we set
$\lambda_{\mathrm{FR}}=\lambda_{\mathrm{TV}}=1.0$ in the full model.

The training loss used class-weighted focal loss~\cite{lin2017focal} with $\gamma{=}2.0$ and
label smoothing of 0.05, with $\lambda_\mathrm{fire}{=}10^{-3}$ and
$\lambda_\mathrm{lasso}{=}10^{-7}$ for the firing-rate and weight
regularizers, respectively. For LIF neurons, we used membrane leak
$\beta{=}0.5$, firing threshold $v_\mathrm{th}{=}1.0$, and sigmoid surrogate
steepness $\alpha_s{=}4.0$.

\subsection{Main Results}
Table~\ref{tab:main_results} summarizes the main comparison.
SCINTILLA-SNN achieves the highest AUROC of 0.748 with 2.83M
parameters. Compared with CNN baselines, SCINTILLA-SNN improves
discrimination while maintaining a compact parameter count. Compared
with transformer baselines, including Vision Transformer and Swin
Transformer, it achieves higher AUROC with lower operation-level energy.
The pure spiking baselines show substantially lower energy consumption
but lower AUROC, suggesting that spike-based efficiency alone is
insufficient for sparse PNI-related patterns. Overall, SCINTILLA-SNN
provides the best accuracy--energy trade-off among the evaluated
models.

\begin{table}[t]
\centering
\caption{Performance and operation-level energy comparison under 5-fold cross-validation.}
\label{tab:main_results}
\resizebox{\linewidth}{!}{
\begin{tabular}{lcccc}
\toprule
Model & Param (M) & OPs (G) & Energy (mJ) & AUROC \\
\midrule
ResNet-18~\cite{he2016deep} & 33.2 & 62.71 & 288.47 & 0.695 \\
DenseNet-121~\cite{huang2017densely} & 11.3 & 10.25 & 47.14 & 0.714 \\
Vision Transformer~\cite{dosovitskiy2020image} & 86.0 & 33.22 & 152.81 & 0.702 \\
Swin Transformer~\cite{liu2021swin} & 27.85 & 17.58 & 80.87 & 0.724 \\
\midrule
Spikformer~\cite{zhou2022spikformer} & 0.87 & 1.46 & 5.40 & 0.649 \\
Spike-driven Transformer~\cite{yao2023spike} & 0.87 & 1.43 & 5.38 & 0.679 \\
\midrule
\textbf{SCINTILLA-SNN (Ours)} & \textbf{2.83} & \textbf{21.92} & \textbf{35.99} & \textbf{0.748} \\
\bottomrule
\end{tabular}
}
\end{table}

\subsection{Qualitative Visualization}

\begin{figure}[t]
\centering
\includegraphics[width=\columnwidth]{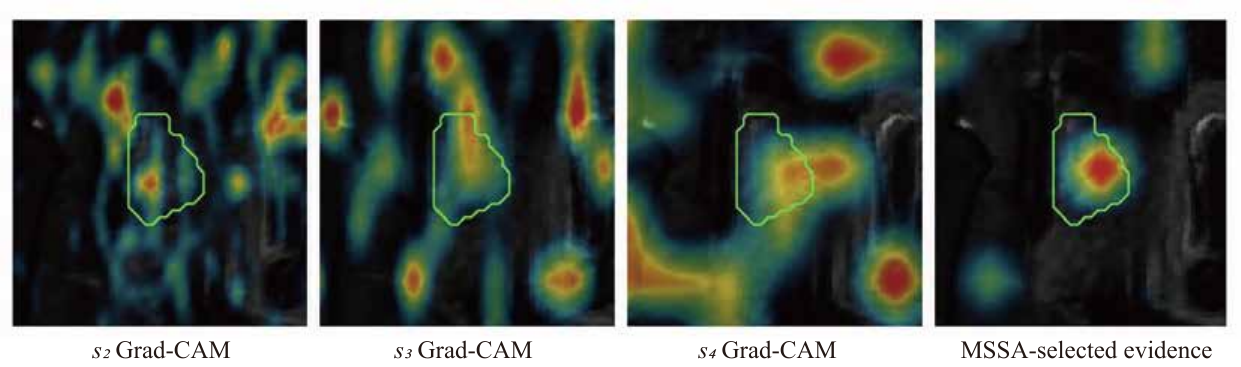}
\caption{Stage-wise Grad-CAM and MSSA-selected evidence for a representative PNI-positive case, overlaid on the T2-weighted image (tumor contour in green).}
\label{fig:gradcam_mssa}
\end{figure}

Fig.~\ref{fig:gradcam_mssa} shows a representative PNI-positive case.
The stage-wise Grad-CAM maps exhibit diffuse responses, whereas the
MSSA-selected evidence is more concentrated near the tumor contour. This
supports the role of MSSA in selecting sparse, tumor-associated tokens
from multi-scale features.

\subsection{Ablation Study}
Table~\ref{tab:ablation} summarizes MSSA ablations under 5-fold
cross-validation. Removing the firing-rate input causes the largest AUROC drop and pushes performance below the content-only baseline, indicating that firing-rate activity is the dominant gating cue and that membrane-potential variability alone is insufficient. In contrast, removing TV causes a smaller drop while remaining above the content-only baseline, and the full model outperforms both single-cue variants, indicating that TV is most useful as a complementary cue to FR and that the two cues jointly refine token selection. Replacing top-$k$ selection with global average pooling further degrades performance, supporting selective aggregation for sparse PNI-related features.

\begin{table}[t]
\centering
\caption{Ablation on MSSA components under 5-fold cross-validation.}
\label{tab:ablation}
\begin{tabular}{lc}
\toprule
Variant & AUROC \\
\midrule
SCINTILLA-SNN & $0.748$ \\
\quad w/o FR input & $0.662$ \\
\quad w/o TV input & $0.721$ \\
\quad content only & $0.704$ \\
\quad top-$k$ $\to$ global average pool & $0.700$ \\
\bottomrule
\end{tabular}
\end{table}

\subsection{Energy Analysis}

Energy consumption is reported using operation-level estimates based on a 45-nm energy model with
$E_{\mathrm{MAC}}=4.6$ pJ and $E_{\mathrm{AC}}=0.9$ pJ~\cite{horowitz20141}. Layers
receiving continuous-valued inputs are counted as MAC operations,
whereas convolutional and linear layers receiving binary spike inputs
are counted as firing-rate-weighted synaptic operations
(SOPs), implemented as AC operations~\cite{zhou2022spikformer,yao2023spike}:
\begin{equation}
\mathrm{SOP}(l)=\bar{s}_l \cdot \Phi_l,
\end{equation}
where $\bar{s}_l$ is the observed input firing rate of layer $l$ and
$\Phi_l$ is the dense operation count of the corresponding layer.

SCINTILLA-SNN also includes auxiliary operations (LIF updates, normalization, element-wise window modulation, fixed scaling, MSSA cues, top-$k$ selection, and softmax pooling). To avoid
underestimating this overhead, we conservatively charge all auxiliary
operations at the MAC energy cost:
\begin{equation}
E_{\mathrm{total}}
=
E_{\mathrm{MAC}}\Phi_{\mathrm{MAC}}
+
E_{\mathrm{AC}}\Phi_{\mathrm{SOP}}
+
E_{\mathrm{MAC}}\Phi_{\mathrm{aux}}.
\end{equation}

For the full SCINTILLA-SNN model, the dense ANN-equivalent
Conv/Linear cost is 181.33G operations, corresponding to 834.10 mJ
if computed entirely as MAC operations. With module-wise firing-rate
profiling, the effective computation consists of 1.82G MACs, 17.53G
AC/SOPs, and 2.58G auxiliary operations. Under the conservative
auxiliary-as-MAC estimate, SCINTILLA-SNN requires 35.99 mJ,
corresponding to a 23.18$\times$ reduction compared with dense
MAC-only computation.

\section{Conclusion}

We introduced SCINTILLA-SNN, which combines hierarchical volumetric spiking feature extraction with MSSA-based selective multi-scale aggregation for MRI-based PNI prediction. By modulating token content with spike-dynamics cues derived from firing rate and membrane-potential variability, the model selectively aggregates diagnostic evidence rather than uniformly pooling volumetric features, achieving the best accuracy–energy trade-off among the evaluated models on a single-center CCA cohort. Going forward, external, multi-institutional validation is needed to assess its generalizability across imaging protocols and clinical settings.

\section*{Acknowledgment}
This work was supported by the IITP grant (IITP-2023-RS-2023-00256081) funded by MSIT, Korea, and the ANCHOR program (2026-ANCHOR-01-110) funded by the Ministry of Education and the Seoul Metropolitan Government, Republic of Korea.

\bibliographystyle{IEEEtran}
\bibliography{ref}
 
\end{document}